%% file: root.tex
\documentclass[letterpaper, 10 pt, conference]{ieeeconf}  

\IEEEoverridecommandlockouts                              

\usepackage{hyperref}
\usepackage{url}
\usepackage{graphicx}
\usepackage{booktabs} 
\usepackage[utf8]{inputenc}
\usepackage{textgreek}
\usepackage{tcolorbox}
\usepackage{bbm}
\usepackage[longend]{algorithm2e} 

\usepackage{wrapfig}
\usepackage{multirow}
\usepackage{float}
\usepackage{subfig}
\input{math_commands.tex}

\title{\Large \bf
Deep Stochastic Kinematic Models for 
Probabilistic Motion Forecasting in Traffic
}

\author{Laura Zheng$^1$, Sanghyun Son$^1$, Jing Liang$^1$, Xijun Wang$^1$, Brian Clipp$^2$, and Ming C. Lin$^1$ \\
    \thanks{The authors are with (1)
 Department of Computer Science, 
         University of Maryland at College Park, MD, U.S.A. and (2) Kitware.
        E-mail: \{lyzheng,shh1295,jingl,xijun,lin\}@umd.edu, brian.clipp@kitware.com}
       \href{https://gamma.umd.edu/sktraj}{\texttt{gamma.umd.edu/sktraj}}
       }

\begin{document}

\maketitle
\thispagestyle{empty}
\pagestyle{empty}

\begin{abstract}
In trajectory forecasting tasks for traffic, future output trajectories can be computed by advancing the ego vehicle's state with predicted actions according to a kinematics model.
By unrolling predicted trajectories via time integration and models of kinematic dynamics, predicted trajectories should not only be kinematically feasible but also relate uncertainty from one timestep to the next. 
While current works in probabilistic prediction do incorporate kinematic priors for mean trajectory prediction, \textit{variance} is often left as a learnable parameter, despite uncertainty in one time step being inextricably tied to uncertainty in the previous time step.
In this paper, we show simple and differentiable analytical approximations describing the relationship between variance at one timestep and that at the next with the kinematic bicycle model.
In our results, we find that encoding the relationship between variance across timesteps works especially well in unoptimal settings, such as with small or noisy datasets.
We observe up to a 50\% performance boost in partial dataset settings and up to an 8\% performance boost in large-scale learning compared to previous kinematic prediction methods on SOTA trajectory forecasting architectures out-of-the-box, with no fine-tuning.

\end{abstract}


\input{sections/intro}
\input{sections/related}
\input{sections/prelim}
\input{sections/method}

\input{sections/results}
\input{sections/conclusion}

\bibliographystyle{ieeetr}
\bibliography{references}

\input{sections/appendix}

\end{document}

%% file: math_commands.tex
\usepackage{amsmath,amsfonts,bm}

\def\eqref#1{equation~\ref{#1}}

\def\1{\bm{1}}

\def\vv{{\bm{v}}}

\DeclareMathAlphabet{\mathsfit}{\encodingdefault}{\sfdefault}{m}{sl}
\SetMathAlphabet{\mathsfit}{bold}{\encodingdefault}{\sfdefault}{bx}{n}



%% file: sections/intro.tex
\section{Introduction}

Motion forecasting in traffic involves predicting the next several seconds of movement for select actors in a scene, given the context of historical trajectories and the local environment.
There are several different approaches to motion forecasting in deep learning, but one thing is in common---scaling resources, dataset sizes, and model sizes will always help generalization capabilities. 
However, the difficult part of motion forecasting is not necessarily \textit{how} the vehicles move, but rather, figuring out the \textit{why}, as there is always a human (at least, for now) behind the steering wheel.

\begin{figure}
    \centering
    \includegraphics[width=0.45\textwidth]{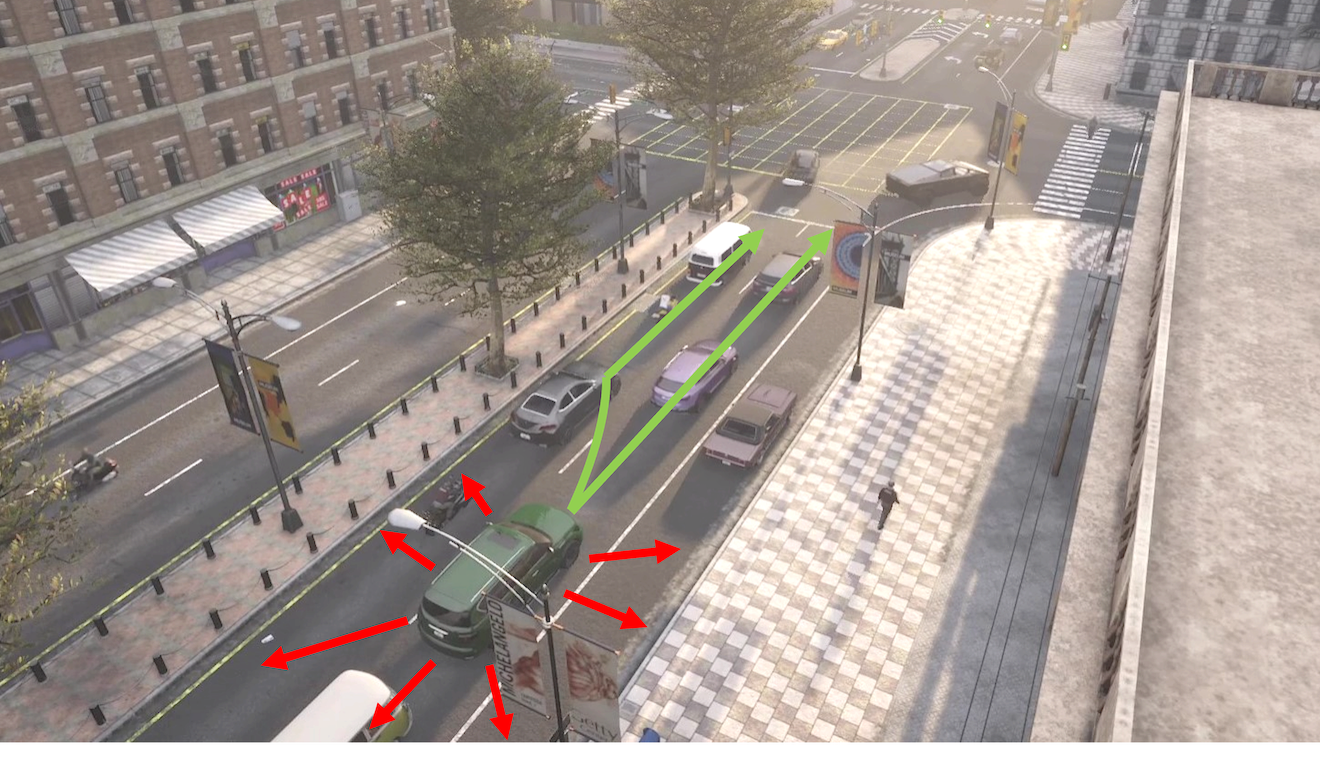}
    \vspace*{-0.5em}
    \caption{\textbf{Motivating example for probabilistic kinematic priors. } In real-world traffic, vehicles have a constrained range of behaviors. For instance, it is not possible for a vehicle to move side to side directly, and driving in reverse is highly unlikely. Without kinematic priors, neural networks may search the space of all trajectories, possible or impossible. Without accounting for analytical variances in trajectories as in previous work, the range of possible future trajectories may also be unrealistic.
    }
    \vspace{-1.5em}
    \label{fig:teaser}
\end{figure}

As model sizes become bigger and performance increases, so does the complexity of learning the basics of vehicle dynamics, where, most of the time, vehicles travel in relatively straight lines behind the vehicle leading directly in front.
Since training a deep neural network is costly, it is beneficial for both resource consumption and model generalization to incorporate the use of existing dynamics models in the training process, combining existing knowledge with powerful modern architectures. 
With embedded priors describing the relationships between variables across time, perhaps a network's modeling power can be better allocated to understanding the \textit{why's} of human behavior, rather than the \textit{how's}.

Kinematic models have already been widely used in various autonomous driving tasks, especially trajectory forecasting and simulation tasks~\cite{Cui_Nguyen_Chou_Lin_Schneider_Bradley_Djuric_2020, Anderson_Vasudevan_Johnson-Roberson_2021, Suo_Wong_Xu_Tu_Cui_Casas_Urtasun_2023, varadarajan2022multipath++}. 
These models explicitly describe how changes in the input parameters influence the output of the dynamical system. 
Typically, kinematic input parameters are often provided by either the robot policy as an action, or by a human in direct interaction with the robot, e.g. steering, throttle, and brake for driving a vehicle. 
For tasks modeling decision-making, such as trajectory forecasting of traffic agents, modeling the input parameters may be more descriptive and interpretable than modeling the output directly. 
Moreover, kinematic models relate the input actions directly to the output observation; thus, any output of the kinematic model should, at the very least, be physically feasible in the real world~\cite{Cui_Nguyen_Chou_Lin_Schneider_Bradley_Djuric_2020}.

\begin{figure*}
  \centering
  \includegraphics[width=0.9\textwidth]{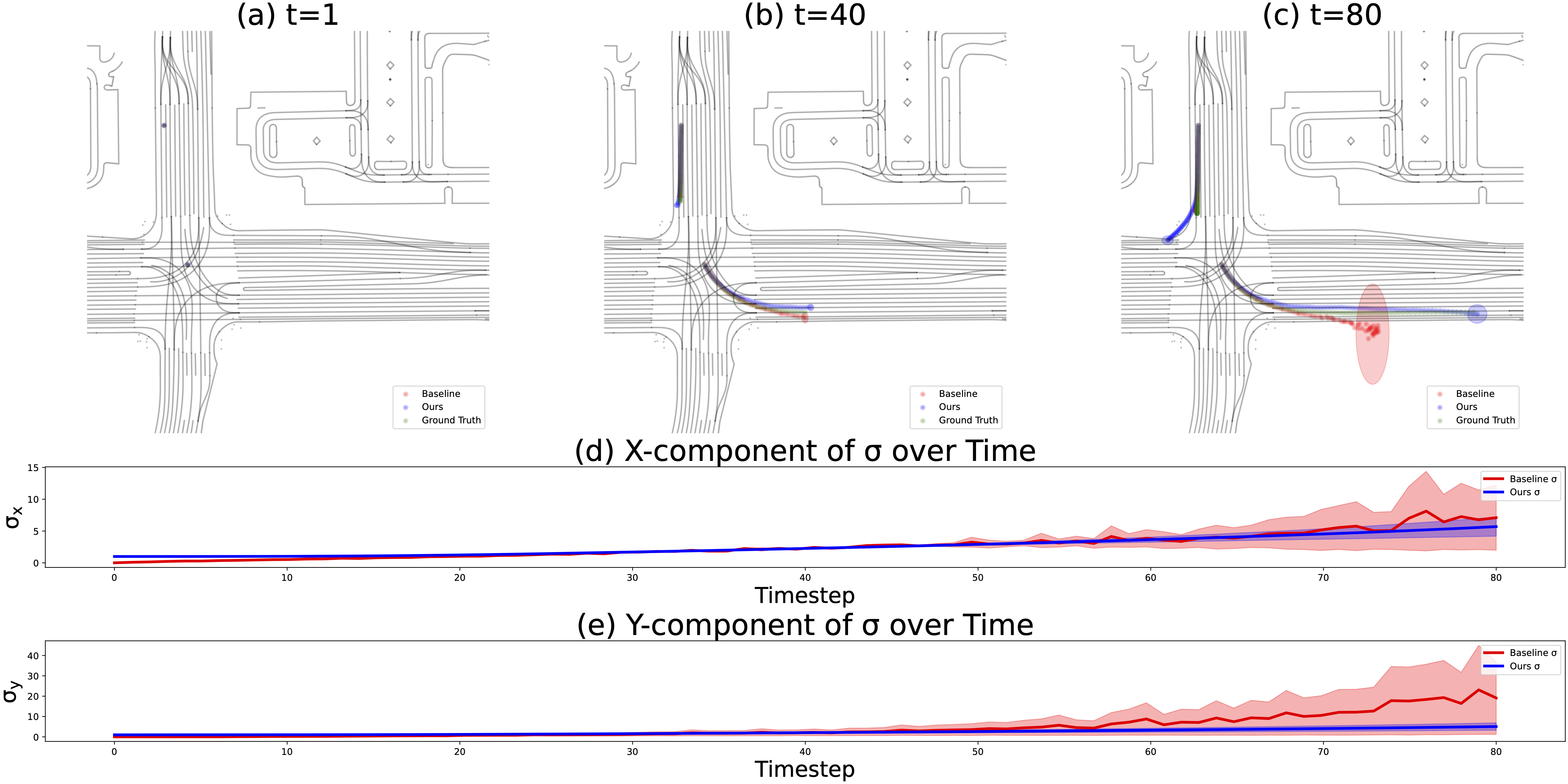}
  \caption{\textbf{Qualitative example.} We visualize an example of the mean trajectory of the highest-scored Gaussian from the results of Table~\ref{tb:1p_waymo}. Standard deviations are visualized as ellipses at each second into the future, and ground truth trajectories are drawn in green. Our method (blue) not only predicts a smoother, less jagged mean trajectory but also provides a more realistic spread of trajectories into the future. The baseline method (red) shows uneven speeds (ellipses are at uneven intervals) in addition to uncertainty extending far beyond the road boundaries.
  } \label{fig:qualitative}
  \vspace*{-1.5em}
\end{figure*}

While previous works already use kinematic models to provide feasible output trajectories, none consider compounding uncertainty into the time horizon analytically. 
Previous works apply kinematic models deterministically, either for deterministic trajectory prediction or to predict the average trajectory in probabilistic settings, leaving the variance as a learnable matrix.

Kalman filtering is a relevant and classical technique that uses kinematic models for trajectory prediction. 
Kalman filters excel in time-dependent settings where signals may be contaminated by noise, such as object tracking and motion prediction~\cite{FARAHI2020115751, Prevost_Desbiens_Gagnon_2007}.
Unlike classical Kalman filtering methods for motion prediction, however, we focus on models of {\em learnable actions and behaviors}, rather than constant models of fixed acceleration and velocity. 
We believe that, with elements from classical methods such as Kalman filtering in addition to powerful modern deep learning architectures, motion prediction in traffic can have the best of both worlds where modeling capability meets stability.

In this paper, we go one step further from previous work by considering the relationship between distributions of kinematic parameters to distributions of trajectory rollouts, rather than single deterministic kinematic parameter predictions to the mean of trajectory rollouts~\cite{Cui_Nguyen_Chou_Lin_Schneider_Bradley_Djuric_2020, varadarajan2022multipath++}. 
We hypothesize that modeling uncertainty explicitly across timesteps according to a kinematic model will result in better performance, realistic trajectories, and stable learning, especially in disadvantageous settings such as small or noisy datasets, which can be common settings, especially in fine-tuning.
We run experiments on four different kinematic formulations with Motion Transformer (MTR)~\cite{shi2022motion} and observe up to a \textit{50\% performance boost in partial-dataset settings} and up to an \textit{8\% performance boost in full dataset settings} on mean average precision (mAP) metrics for the Waymo Open Motion Dataset (WOMD)~\cite{Ettinger_2021_ICCV} compared to with kinematic priors enforced on mean trajectories only.

In summary, the main contributions of this work include: 

\begin{enumerate} 
    \item A simple and effective method for incorporating analytically-derived kinematic priors into probabilistic models for trajectory forecasting (Section~\ref{sec:method}), which boosts performance and generalization (Section~\ref{sec:results}) and requires trivial additional overhead computation;
    \item Results and analysis in different settings on four different kinematic formulations: velocity components $v_x$ and $v_y$, acceleration components $a_x$ and $a_y$, speed $s$ and heading $\theta$ components, and steering $\delta$ and acceleration $a$ (Section~\ref{sec:kinematics}).
    \item Analytical error bounds for the first and second-order kinematic formulations (section~\ref{sec:error}).
\end{enumerate}

%% file: sections/related.tex
\section{Related Works}

\begin{figure}
    \centering
    \includegraphics[width=0.65\linewidth]{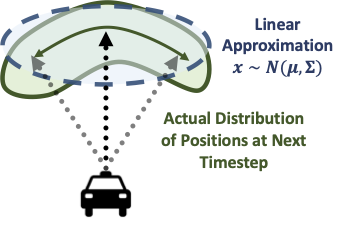}
    \vspace*{-1em}
    \caption{Linear approximation of position distributions via kinematic priors.}
    \vspace*{-1em}
    \label{fig:linear-approximation}
\end{figure}

\subsection{Trajectory Forecasting for Traffic}
Traffic trajectory forecasting is a popular task where the goal is to predict the short-term future trajectory of multiple agents in a traffic scene. Being able to predict the future positions and intents of each vehicle provides context for other modules in autonomous driving, such as path planning. 
Large, robust benchmarks such as the Waymo Motion Dataset~\cite{Ettinger_2021_ICCV, Kan_2023_arxiv}, Argoverse~\cite{argoverse}, and the NuScenes Dataset~\cite{nuscenes} have provided a standardized setting for advancements in the task, with leaderboards showing clear rankings for state-of-the-art models. 
Amongst the top performing architectures, most are based on Transformers for feature extraction~\cite{shi2022motion, qian20232nd, liu2021multimodal, zhou2023query-qcnet, ngiam2021scene}. 
Current SOTA models also model trajectory prediction probabilistically, as inspired by the use of GMMs in MultiPath~\cite{pmlr-v100-chai20a}. 

One common theme amongst relevant state-of-the-art, however, is that works employing kinematic models for time-integrated trajectory rollouts typically only consider kinematic variables deterministically, \textit{which neglect the relation between kinematic input uncertainty and trajectory rollout uncertainty}~\cite{Cui_Nguyen_Chou_Lin_Schneider_Bradley_Djuric_2020, varadarajan2022multipath++, Ścibior_Lioutas_Reda_Bateni_Wood_2021}. 
In our work, we present a method for use of kinematic priors which can be complemented with any previous work in trajectory forecasting. 
Our contribution can be implemented in any of the SOTA methods above, since it is a simple reformulation of the task with no additional information needed. 

\subsection{Physics-based Priors for Learning}
Model-based learning has shown to be effective in many applications, especially in robotics and graphics. There are generally two approaches to using models of the real world: 1) learning a model of dynamics via a separate neural network~\cite{rempe2022strive, lutter2019deep, NEURIPS2019_26cd8eca,janner2019mbpo}, or 2) using existing models of the real world via differentiable simulation~\cite{de2018end, degrave2019differentiable, geilinger2020add, Qiao2020Scalable, son2023gradppo, xu2021accelerated, liang2019differentiable}. 

In our method, we pursue the latter. Since we are not modeling complex systems such as cloth or fluid, simulation of traffic agent states require only a simple, fast, and differentiable update. In addition, since the kinematic models do not describe interactions between agents, the complexity of the necessary model is greatly reduced. In this paper, we hypothesize that modeling the simple kinematics (e.g., how a vehicle moves forward) with equations will allow for greater modeling expressivity on behavior.

%% file: sections/prelim.tex
\section{Kinematics of Traffic Agents}
Our method borrows concepts from simulation and kinematics. 
In a traffic simulation, each vehicle holds some sort of state consisting of position along the global x-axis $x$, position along the global y-axis $y$, velocity $v$, and heading $\theta$. 
This state is propagated forward in time via a kinematic model describing the constraints of movement with respect to some input acceleration $a$ and steering angle $\delta$ actions. 
The Bicycle Model, both classical and popularly utilized in path planning for robots, describes the kinematic dynamics of a wheeled agent given its length $L$: 

\begin{align*}
    \frac{d}{dt} 
    \begin{pmatrix}
        x \\
        y \\ 
        \theta \\
        v 
    \end{pmatrix} = 
    \begin{pmatrix}
        \dot{x} \\
        \dot{y} \\ 
        \dot{\theta} \\
        \dot{v} 
    \end{pmatrix} = 
    \begin{pmatrix}
        v \cdot cos\theta \\
        v \cdot sin\theta \\
        \frac{v \cdot tan\theta}{L} \\
        a
    \end{pmatrix}
\end{align*}

We refer to this model throughout this paper to derive the relationship between predicted distributions of kinematic variables and the corresponding distributions of positions $x$ and $y$ for the objective trajectory prediction task. 

We use Euler time integration in forward simulation of the kinematic model to obtain future positions. As we will show later, explicit Euler time integration is simple for handling Gaussian distributions, despite being less precise than higher-order methods like Runge-Kutta. Furthermore, higher-order methods are also difficult to implement in a parallelizable and differentiable fashion, which would add additional overhead as a tradeoff for better accuracy. 
An agent's state is propagated forward from timestep $t$ to $t+1$ with the following, given a timestep interval $\Delta t$: 

\begin{align*}
    \begin{pmatrix}
        x_{t+1} \\
        y_{t+1} \\ 
        \theta_{t+1} \\
        v_{t+1} 
    \end{pmatrix} 
    =
    \begin{pmatrix}
        x_{t} \\
        y_{t} \\ 
        \theta_{t} \\
        v_{t} 
    \end{pmatrix}
    + 
    \begin{pmatrix}
        \dot{x} \\
        \dot{y} \\ 
        \dot{\theta} \\
        \dot{v} 
    \end{pmatrix} \cdot \Delta t 
\end{align*}


%% file: sections/method.tex
\section{Methodology}
\label{sec:method}
\subsection{Probabilistic Trajectory Forecasting}
Trajectory forecasting is a popular task in autonomous systems where the objective is to predict the future trajectory of multiple agents for $T$ total future timesteps, given a short trajectory history. 
Recently, state-of-the-art methods~\cite{multipath, varadarajan_multipathplusplus22, wang2023multiverse, shi2022motion} utilize Gaussian Mixture Models (GMMs) to model the distribution of potential future trajectories, given some intention waypoint or destination of the agent and various extracted agent or map features.
Each method utilizes GMMs slightly differently, however, all methods use GMMs to model the distribution of future agent trajectories. 
We apply a kinematic prior to the GMM head directly---thus, our method is agnostic to the design of the learning framework.
Instead of predicting a future trajectory deterministically, current works instead predict a mixture of Gaussian components $\left(\mu_x, \mu_y, \sigma_x, \sigma_y, \rho \right)$ describing the mean $\mu$ and standard deviation $\sigma$ of $x$ and $y$, in addition to a correlation coefficient $\rho$ and Gaussian component probability $p$. 
The standard deviation terms, $\sigma_x$ and $\sigma_y$, along with correlation coefficient $\rho$, parameterize the covariance matrix of a Gaussian centered around $\mu_x$ and $\mu_y$.

Ultimately, the prediction objective is, for each timestep, to maximize the log-likelihood of the ground truth trajectory waypoint $(x, y)$ belonging to the position distribution outputted by the GMM: 
\begin{align*}
    \mathcal{L} = - \log p_h - \log{\mathcal{N}_h(x - \hat{\mu}_x, \hat{\sigma}_x; y - \hat{\mu}_y; \rho)}
\end{align*}

This formulation assumes that distributions between timesteps are conditionally independent, similarly to Multipath~\cite{multipath} and its derivatives. 
Alternatively, it's possible to implement predictions with GMMs in an autoregressive manner, where trajectory distributions are dependent on the position of the previous timestep. 
The drawback of this is the additional overhead of computing conditional distributions with recurrent architectures, rather than jointly predicting for all timesteps at once. 


\subsection{Kinematic Priors in Gaussian-Mixture Model Predictions}
\label{sec:kinematics}

The high-level idea for kinematic priors is simple: instead of predicting the distribution of positions at each timestep, we can instead predict the distribution of first-order or second-order kinematic terms and then use time integration to derive the subsequent position distributions. 

The intuition for enforcing kinematic priors comes from the idea that even conditionally independent predicted trajectory waypoints have inherent relationships with each other depending on the state of the agent, even if the neural network does not model it. By propagating these relationships across the time horizon, we focus optimization of the network in the space of kinematically feasible trajectories. 
We consider four different formulations: 1) with velocity components $v_x$ and $v_y$, 2) with acceleration components $a_x$ and $a_y$, 3) with speed $s = \|\vv\|$ and heading $\theta$, and finally, 4) with acceleration $a$ (second order of speed) and steering angle $\delta$. 


\subsubsection{Formulation 1: Velocity Components}
\label{sec:formulation1-velxvely}
\input{sections/formulations/formulation1}


\subsubsection{Formulation 2: Acceleration Components}
\label{sec:formulation2-accelxaccely}
\input{sections/formulations/formulation2}

\subsubsection{Formulation 3: Speed and Heading}
\label{sec:formulation3-speed-heading}
\input{sections/formulations/formulation3}

\subsubsection{Formulation 4: Acceleration and Steering}
\label{sec:formulation4-accel-steering}
\input{sections/formulations/formulation4}

\subsection{Error Bound of Linear Approximation.} 
\label{sec:error}
\input{sections/formulations/error_bound}

%% file: sections/formulations/formulation1.tex
The velocity component formulation is the simplest kinematic formulation, where the GMMs predict the distribution of velocity components $v_x$ and $v_y$ for each timestep $t$.

Our goal is to derive $(\mu^{t+1}_x,\mu^{t+1}_y,\sigma^{t+1}_x,\sigma^{t+1}_y)$ given $(\mu^{t}_{v_x}, \mu^{t}_{v_y}, \sigma^{t}_{v_x}, \sigma^{t}_{v_y})$. 
In the deterministic setting, the position at the next timestep can be generated via Euler time integration given a time interval (in seconds) $\Delta t$, which varies depending on the dataset: 
\begin{align*}
    x^{t+1} &= x^t + v^t_x \cdot \Delta t
\end{align*}
If we consider both $x$ and $v$ to be Gaussian distributions rather than scalar values, we can represent the above in terms of distribution parameters below with the reparameterization trick used in Variational Autoencoders (VAEs)~\cite{kingma2022autoencoding}: 
\begin{align*}
    \mathcal{N}_{x^{t+1}} &= (\mu^t_{x} + \sigma^t_{x} \cdot \epsilon_x) + (\mu^t_{v_x} + \sigma^t_{v_x} \cdot \epsilon_v) \cdot \Delta t
\end{align*}
where $\epsilon_x, \epsilon_v \sim \mathcal{N}(0, 1)$.
By grouping deterministic (without $\epsilon$) and probabilistic terms (with $\epsilon$), we obtain the reparameterized form of the Gaussian distribution describing $x^{t+1}$:
\begin{align*}
    \mathcal{N}_{x^{t+1}} &= (\mu^t_{x} + \mu^t_{v_x} \cdot \Delta t) + (\sigma^t_{x} \cdot \epsilon_x + \sigma^t_{v_x} \cdot \Delta t \cdot \epsilon_v) \\
\end{align*}
\vspace{-1em}
and thus,
\begin{align}
\label{eq:formulation1}
    \mu^{t+1}_{x} &= \mu^t_x + \mu^t_{v_x} \cdot \Delta t , \
    \sigma^{t+1}_{x} = \sqrt{{\sigma^t_{x}}^2 + {\sigma^t_{v_x}}^2 \cdot \Delta t^2} \\
    \mu^{t+1}_{y} &= \mu^t_y + \mu^t_{v_y} \cdot \Delta t , \
    \sigma^{t+1}_{y} = \sqrt{{\sigma^t_{y}}^2 + {\sigma^t_{v_y}}^2 \cdot \Delta t^2}
\end{align}
Also, for the first prediction timestep, we consider the starting trajectory position to represent a distribution with standard deviation equal to zero.

This Gaussian form is also intuitive as the sum of two Gaussian random variables is also Gaussian. 
Since this formulation is not dependent on any term outside of timestep $t$, distributions for all $T$ timesteps can be computed with vectorized cumulative sum operations. 
We derive the same distributions in the following sections in a similar fashion with different kinematic parameterizations. 

\begin{figure}[t!]
    \centering
    \vspace*{-0.5em}
    \includegraphics[width=0.45\textwidth]{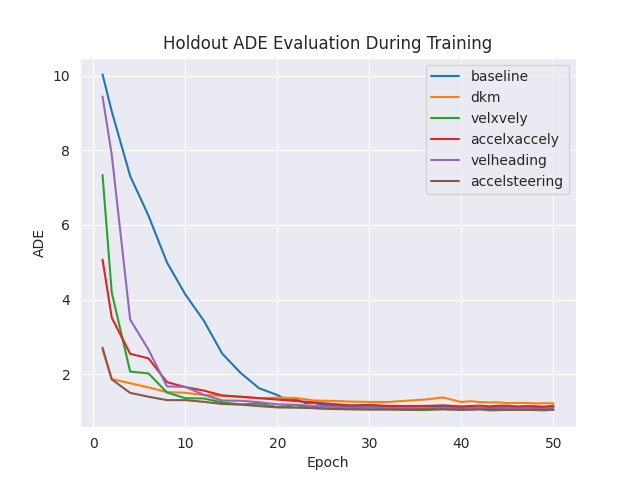}
    \vspace*{-0.5em}
    \caption{\textbf{Average Displacement Error (ADE) across training across each method and formulation in the small dataset setting.} Learning with stochastic kinematic priors aids with faster learning. Compared to the baseline (blue), all models employing kinematic priors converge much more quickly. Formulation 4 (brown) converges most quickly.}
    \vspace{-1em}
    \label{fig:ade_training_curve}
\end{figure}


%% file: sections/formulations/formulation2.tex
Following directly from the first formulation above, we now consider the second-order case where the GMM predicts acceleration components. 
Here, our goal is to derive $(\mu^{t+1}_x,\mu^{t+1}_y,\sigma^{t+1}_x,\sigma^{t+1}_y)$ given $(\mu^{t}_{a_x}, \mu^{t}_{a_y}, \sigma^{t}_{a_x}, \sigma^{t}_{a_y})$. 
The deterministic relationship between acceleration components $a_x$ and $a_y$ with $v_x$ and $v_y$ via Euler time integration is simply 
\begin{align*}
    v^{t+1}_x = v^t_x + a^t_x \cdot \Delta t \\
    v^{t+1}_y = v^t_y + a^t_y \cdot \Delta t
\end{align*}

Following similar steps to Formulation 1, we obtain the parameterized distributions of $v_x$ and $v_y$: 

\begin{align}
\label{eq:formulation2}
    \mu^{t+1}_{v_x} &= \mu^t_{v_x} + \mu^t_{a_x} \cdot \Delta t , \
    \sigma^{t+1}_{v_x} = \sqrt{{\sigma^t_{v_x}}^2 + {\sigma^t_{a_x}}^2 \cdot \Delta t^2} \\
    \mu^{t+1}_{v_y} &= \mu^t_{v_y} + \mu^t_{a_y} \cdot \Delta t , \
    \sigma^{t+1}_{v_y} = \sqrt{{\sigma^t_{v_y}}^2 + {\sigma^t_{a_y}}^2 \cdot \Delta t^2}
\end{align}

When computed for all timesteps, we now have $T$ total distributions representing $v_x$ and $v_y$, which then degenerates to Formulation 1 in Equation~\ref{eq:formulation1}.

%% file: sections/formulations/formulation3.tex
Now, we derive the approximated analytical form of position distributions according to first-order dynamics of the Bicycle Model, speed $s = \|\vv\|$ and heading $\theta$.
Similarly as before, our goal is to derive $(\mu^{t+1}_x,\mu^{t+1}_y,\sigma^{t+1}_x,\sigma^{t+1}_y)$ given $(\mu^{t}_{s}, \mu^{t}_{\theta}, \sigma^{t}_{s}, \sigma^{t}_{\theta})$. 
Deterministically, we can get the update for $x^{t+1}$ and $y^{t+1}$ with the following relation from the Bicycle Model: 
\begin{align*}
    \begin{bmatrix}
        x^{t+1} \\
        y^{t+1}
    \end{bmatrix} = 
    \begin{bmatrix}
        x^t + s^t \cdot \cos \theta \cdot \Delta t \\
        y^t + s^t \cdot \sin \theta \cdot \Delta t
    \end{bmatrix}
\end{align*}
When representing this formulation in terms of Gaussian parameters, we point out that the functions $cos(\cdot)$ and $sin(\cdot)$ applied on Gaussian random variables do not produce Gaussians: 
\begin{align*}
\resizebox{.45 \textwidth}{!}{$
    \begin{bmatrix}
        \mathcal{N}_x^{t+1} \\
        \mathcal{N}_y^{t+1}
    \end{bmatrix} \neq 
    \begin{bmatrix}
        (\mu^t_x + \sigma^t_s \cdot \epsilon_x) + (\mu^t_{s} + \sigma^t_s \cdot \epsilon_s) \cdot \cos (\mu^t_\theta + \sigma^t_\theta \cdot \epsilon_x) \cdot \Delta t \\
        (\mu^t_y + \sigma^t_s \cdot \epsilon_y) + (\mu^t_{s} + \sigma^t_s \cdot \epsilon_s) \cdot \sin (\mu^t_\theta + \sigma^t_\theta \cdot \epsilon_y) \cdot \Delta t
    \end{bmatrix}$}
\end{align*}

To amend this, we instead replace $cos(\cdot)$ and $sin(\cdot)$ with linear approximations $T(\cdot)$ evaluated at $\mu_\theta$. 
\begin{align*}
    T_{\sin}(\theta) = \sin (\mu_\theta) + \cos(\mu_\theta) \cdot (\theta - \mu_\theta) \\
    T_{\cos}(\theta) = \cos(\mu_\theta) - \sin(\mu_\theta) \cdot (\theta - \mu_\theta)
\end{align*}
We now derive the formulation of the distribution of positions with the linear approximations instead:

\begin{align*}
    \begin{bmatrix}
        \mathcal{N}_{x^{t+1}} \\
        \mathcal{N}_{y^{t+1}}
    \end{bmatrix} &\approx
    \begin{bmatrix}
        \mu_{x}^{t+1} + \sigma_{x}^{t+1} \cdot \epsilon_x \\
        \mu_{y}^{t+1} + \sigma_{y}^{t+1} \cdot \epsilon_y
    \end{bmatrix} \\
\end{align*}

where 

\begin{align}
\label{eq:formulation3}
    \mu_x^{t+1} &= \mu_x^t + \mu_s^t \cdot cos(\mu_\theta) \cdot \Delta t \\
    \sigma_x^{t+1} &= \sqrt{{\sigma_x^t}^2 + A^2 + B^2 + C^2} \\
    A &= \mu^t_s \cdot \sigma_\theta^t \cdot sin(\mu_\theta^t) \cdot \Delta t \\
    B &= \sigma_s^t \cdot cos(\mu_\theta^t) \cdot \Delta t \\
    C &= \sigma_s^t \cdot \sigma_\theta^t \cdot sin(\mu_\theta^t) \cdot \Delta t 
\end{align}
and 
\begin{align}
    \mu_y^{t+1} &= \mu_y^t + \mu_s^t \cdot sin(\mu_\theta) \cdot \Delta t \\
    \sigma_y^{t+1} &= \sqrt{{\sigma_y^t}^2 + D^2 + E^2 + F^2} \\
    D &= \mu^t_s \cdot \sigma_\theta^t \cdot cos(\mu_\theta^t) \cdot \Delta t \\
    E &= \sigma_s^t \cdot sin(\mu_\theta^t) \cdot \Delta t \\
    F &= \sigma_s^t \cdot \sigma_\theta^t \cdot cos(\mu_\theta^t) \cdot \Delta t 
\end{align}

The full expansion of these terms is described in Section~\ref{sec:appendix_formulation3_expansion} of the appendix, which can be found on our project page under the title. 




%% file: sections/formulations/formulation4.tex
Lastly, we derive a second-order kinematic formulation based on the bicycle model: steering $\delta$ and acceleration $a$. This formulation is the second-order version of the velocity-heading formulation. Here, we assume acceleration $a$ to be scalar and directionless, in contrast to Formulation 2, where we consider acceleration to be a vector with lateral and longitudinal components. Similarly to Formulation 2, we also use the linear approximation of $\tan(\cdot)$ in order to derive approximated position distributions. 

Following the Bicycle Model, the update for speed and heading at each timestep is: 

\begin{align*}
    \begin{bmatrix}
        s^{t+1} \\
        \theta^{t+1}
    \end{bmatrix} = 
    \begin{bmatrix}
        s^t + a \cdot \Delta t \\
        \theta^t + \frac{s \cdot \tan(\delta)}{L} \cdot \Delta t
    \end{bmatrix}
\end{align*}

Where $L$ is the length of the agent. 
When we represent this process probabilistically as random Gaussian variables, we solve for the distributions of first-order variables, speed ($s$) and heading ($\theta$): 

\begin{align*}
    \begin{bmatrix}
        \mathcal{N}_{s^{t+1}} \\
        \mathcal{N}_{\theta^{t+1}}
    \end{bmatrix} &\approx
    \begin{bmatrix}
        \mu_{\theta}^{t+1} + \sigma_{s}^{t+1} \cdot \epsilon_s \\
        \mu_{\theta}^{t+1} + \sigma_{\theta}^{t+1} \cdot \epsilon_\theta
    \end{bmatrix} \\
\end{align*}

where 

\begin{align}
\label{eq:formulation4}
    \mu_s^{t+1} &= \mu_s^t + \mu_a^t \cdot \Delta t \\
    \sigma_s^{t+1} &= \sigma_{s}^t + \sigma_a^{t} \cdot \Delta t \\
    \mu_\theta^{t+1} &= \mu_\theta^t + \frac{1}{L} \cdot (\mu_s^t \cdot tan(\mu_\delta^t)) \cdot \Delta t \\
    \sigma_\theta^{t+1} &= \sqrt{{\sigma_\theta^t}^2 + X^2 + Y^2 + Z^2} \\
    X &= \frac{1}{L} \cdot \mu_s^t \cdot \sigma_\delta^t \cdot \frac{1}{cos^2(\mu_\delta^t)} \cdot \Delta t \\
    Y &= \frac{1}{L} \cdot \sigma_s^t \cdot tan(\mu_\delta^t) \cdot \Delta t \\
    Z &= \frac{1}{L} \cdot \sigma_s^t \cdot \sigma_\delta^t \cdot \frac{1}{cos^2(\mu_\delta^t)} \cdot \Delta t 
\end{align}

With the computed distributions of speed and heading, we can then use the analytical definitions from Formulation 3, in Equation~\ref{eq:formulation3}, to derive the distribution of positions in terms of $x$ and $y$.

%% file: sections/formulations/error_bound.tex
The linear approximations used to derive the variance in predicted trajectories come with some error relative to the actual variance computed from transformations on predicted Gaussian variables, which may not necessarily be easily computable or known. 
However, we can bound the error of this linear approximation thanks to the alternating property of the sine and cosine Taylor expansions. 

We analytically derive the error bound for the linear approximation for $f(x) = \cos(x)$ and $f(x) = \sin(x)$ functions at $\mu_\theta$. 
Since the Taylor series expansion of both functions are alternating, the error is bounded by the term representing the second order derivative: 
\begin{align*}
    R^{\cos}_2(\mu_\theta + \sigma_\theta \cdot \epsilon_\theta) &\leq \left|\frac{-\cos(\mu_\theta)}{2!}\right| (\mu_\theta + \sigma_\theta \cdot \epsilon_\theta - \mu_\theta)^2 \\
    &= \left|\frac{-\cos(\mu_\theta)}{2}\right| \sigma_\theta^2 \cdot \epsilon_\theta^2 \leq \frac{\sigma_\theta^2}{2}\cdot \epsilon_\theta^2 \\
    R^{\sin}_2(\mu_\theta + \sigma_\theta \cdot \epsilon_\theta) &\leq \left|\frac{-\sin(\mu_\theta)}{2!}\right| (\mu_\theta + \sigma_\theta \cdot \epsilon_\theta - \mu_\theta)^2 \\
    &= \left|\frac{-\sin(\mu_\theta)}{2}\right| \sigma_\theta^2 \cdot \epsilon_\theta^2 \leq \frac{\sigma_\theta^2}{2}\cdot \epsilon_\theta^2
\end{align*}
For both functions, the Lagrange error $R(x) = f(x) - T(x)$ is bounded by $R(x) \leq \frac{\sigma^2_\theta}{2} \cdot \epsilon^2_\theta \leq \frac{\sigma^2_\theta}{2} \cdot \epsilon_\theta^2$. 

In other words, the error from each linear approximation is, at most, on the order of $\sigma_\theta^2 \cdot \epsilon_\theta^2$, where $\epsilon_\theta$ is a random variable sampled from $\mathcal{N}(0, 1)$. 
While this error can compound across timesteps, we point out that $\sigma$ values empirically remain quite small within the task (bounded most of the time by road width or differences in speed), where one unit value is one meter, as well as the random chi-square variable $\epsilon_\theta^2$ being heavily biased towards 0 ($\sim 68\%$ probability under 1, and $\sim 99.5\%$ probability under 4 for normal chi-square distribution).

%% file: sections/results.tex
\section{Results}
\label{sec:results}

In this section, we show experiments that highlight the effect of kinematic priors on performance. We implement kinematic priors on state-of-the-art method Motion Transformer (MTR)~\cite{shi2022motion}, which serves as our baseline method.  

\textbf{Hardware.} We train all experiments on eight RTX A5000 GPUs, with 64 GB of memory and 32 CPU cores. Experiments on the full dataset are trained for 30 epochs, while experiments with the smaller dataset are trained for 50 epochs. Additionally, we downscale the model from its original size of 65 million parameters to 2 million parameters and re-train all models under these settings for fair comparison. 
Furthermore, we re-implement Deep Kinematic Models (DKM) ~\cite{Cui_Nguyen_Chou_Lin_Schneider_Bradley_Djuric_2020} to contextualize our probabilistic method against deterministic methods, as the official implementation is not publicly available. DKM is implemented against the same backbone as vanilla MTR and our models. 
More details on training hyperparameters can be found in Table~\ref{tab:appendix-model-architecture-hyperparamters} of the appendix, which can also be found on the project website. 

\begin{table}[t!]
 \caption{{\bf Performance comparison for vehicles on each kinematic formulation versus SOTA baseline~\cite{shi2022motion} and DKM~\cite{Cui_Nguyen_Chou_Lin_Schneider_Bradley_Djuric_2020} on Waymo Motion Dataset, Marginal Trajectory Prediction. In our experiments, we downscale the backbone model size from 65M parameters to 2M parameters. From the results, we find that {\em Formulation 3 (speed and heading)} provides the greatest and most consistent boost in performance across most metrics over the baseline that does not include kinematic priors.}}
\label{tb:100p_waymo}
  \centering
  \scalebox{.9}{
  \begin{tabular}{l|rrrr}
    \toprule
   Method & mAP$\uparrow$ & minADE$\downarrow$ & minFDE$\downarrow$ & MissRate$\downarrow$ \\
    \midrule
MTR & 0.3872 & 0.8131 & 1.6700 & 0.1817 \\
DKM$_{Uniform}$ & 0.3613 & 0.8724 & 1.7685 & 0.1994 \\
DKM$_{Learnable}$ & 0.3821 & 0.8150 & 1.6788 & 0.1839 \\
Ours$_{F1}$ & 0.3880 & 0.8241 & 1.6783 & 0.1834 \\
Ours$_{F2}$ & 0.3892 & 0.8217 & 1.6710 & 0.1829 \\
Ours$_{F3}$ & \textbf{0.3914} & \textbf{0.8102} & \textbf{1.6634} & \textbf{0.1792} \\
Ours$_{F4}$ & 0.3910 & 0.8220 & 1.6782 & 0.1830 \\
    \bottomrule
  \end{tabular}}
  \vspace*{-1.5em}
\end{table}

\subsection{Performance on Waymo Motion Prediction Dataset}
We evaluate the baseline model and all kinematic formulations on the Waymo Motion Prediction Dataset~\cite{Ettinger_2021_ICCV}. 
The Waymo dataset consists of over 100,000 segments of traffic, where each scenario contains multiple agents of three classes: vehicles, pedestrians, and cyclists. The data is collected from high-quality, high-resolution sensors that sample traffic states at 10 Hz. 
The objective is, given 1 second of trajectory history for each vehicle, to predict trajectories for the next 8 seconds. 
For simplicity, we use the bicycle kinematic model for all three classes and leave discerning between the three, especially for pedestrians, for future work. 

We evaluate our model's performance on Mean Average Precision (mAP), Minimum Average Displacement Error (minADE), minimum final displacement error (minFDE), and Miss Rate, similarly to ~\cite{Ettinger_2021_ICCV}. We reiterate their definitions below for convenience.

\begin{itemize}
    \item Mean Average Precision (mAP): mAP is computed across all classes of trajectories. The classes include straight, straight-left, straight-right, left, right, left u-turn, right u-turn, and stationary. For each prediction, one true positive is chosen based on the highest confidence trajectory within a defined threshold of the ground truth trajectory, while all other predictions are assigned a false positive. Intuitively, the mAP metric describes prediction precision while accounting for all trajectory class types. This is beneficial especially when there is an imbalance of classes in the dataset (e.g., there may be many more straight-line trajectories in the dataset than there are right u-turns). 
    \item Minimum Average Displacement Error (minADE):  average L2 norm between the ground truth and the closest prediction; $minADE(G) = \min_i \frac{1}{T} \sum^{T}_{t=1} \|\hat{s}^t_G - s^{it}_G \|_2$.
    \item Minimum final displacement error (minFDE): L2 norm between only the positions at the final timestep, $T$; $minFDE(G) = \min_i \|\hat{s}^T_G - s^{iT}_G\|_2$. 
    \item Miss Rate: The number of predictions lying outside a reasonable threshold from the ground truth. The miss rate first describes the ratio of object predictions lying outside a threshold from the ground truth to the total number of agents predicted.
\end{itemize}

We show results for the Waymo Motion dataset in Table~\ref{tb:100p_waymo}, where we compared performance across two baselines, MTR~\cite{shi2022motion} and two variants of DKM~\cite{Cui_Nguyen_Chou_Lin_Schneider_Bradley_Djuric_2020}, one with uniform variance and one with learnable variance (used in Multipath~\cite{multipath}), and all formulations. From these results, we observe the greatest improvement over the baseline with Formulation 3, which involves the first-order velocity and heading components. 
We observe that the benefit of our method in full-scale training settings diminishes. 
This performance gap closing may be due to the computational complexity of the network, the large dataset, and or the long supervised training time out-scaling benefits provided by applying kinematic constraints. 
However, deploying models in the wild may not necessarily have such optimal settings, especially in cases of sparse data or domain transfer.
Thus, we also consider the effects of suboptimal settings for trajectory prediction, as we hypothesize that learning first or second-order terms provides information when data cannot. 

\begin{table}[t!]
 \caption{{\bf Performance comparison on vehicles for each kinematic formulation versus SOTA in a small dataset setting. 
 We train models on 1\% of the Waymo Motion Dataset and use the same full evaluation set as that in Table~\ref{tb:100p_waymo}. We see pronounced improvements in performance metrics in settings with significantly less data available, with a nearly 13\% mAP performance gain over the baseline and nearly 50\% mAP performance gain compared to deterministic kinematic method from DKM~\cite{Cui_Nguyen_Chou_Lin_Schneider_Bradley_Djuric_2020} across most formulations. In the case of Formulation 4, analytical modeling of variance contributes an additional 2.5\% in mAP performance over DKM with Learnable Covariance (DKM$_{Learnable}$), which is employed by MultiPath~\cite{multipath}.}}
 \label{tb:1p_waymo}
  \centering
  \scalebox{.9}{
  \begin{tabular}{l|rrrr}
    \toprule
   Method &  mAP$\uparrow$ &  minADE$\downarrow$ &  minFDE$\downarrow$ &  MissRate$\downarrow$ \\
    \midrule
MTR & 0.1697 & 1.4735 & 3.7602 & 0.3723 \\
DKM$_{Uniform}$ & 0.1283 & 1.6632 & 3.7148 & 0.4361 \\
DKM$_{Learnable}$ & 0.1811 & 1.3082 & 2.9044 & 0.3612 \\
Ours$_{F1}$ & 0.1907 & \textbf{1.2819} & 2.7956 & 0.3489 \\
Ours$_{F2}$ & \textbf{0.1932} & 1.4141 & \textbf{2.6953} & \textbf{0.3376} \\
Ours$_{F3}$ & 0.1833 & 1.3525 & 2.8362 & 0.3470 \\
Ours$_{F4}$ & 0.1859 & 1.2932 & 2.8271 & 0.3502 \\
    \bottomrule
  \end{tabular}}
  \vspace*{-2em}
\end{table}

\subsection{Performance on a Smaller Dataset Setting}
We examine the effects of kinematic priors on a smaller dataset size. This is motivated by imbalance of scenarios in driving datasets, where many samples are representative of longitudinal straight-line driving or stationary movement, and much less are representative of extreme lateral movements such as U-turns. Thus, large and robust benchmarks like the Waymo, Nuscenes, and Argoverse datasets are necessary for learning robust models. 

We train the baseline model and all formulations on only 1\% of the original Waymo dataset and benchmark their performance on 100\% of the evaluation set in Table~\ref{tb:1p_waymo}. All experiments were trained over 50 epochs. \textbf{{\em In the small dataset setting, we observe that providing a kinematic prior in any form improves performance on all metrics except the deterministic case of DKM.} }
Figure~\ref{fig:ade_training_curve} shows how all kinematic formulations improve convergence speed over the baseline, with Formulation 4 (acceleration and steering) converging most quickly.
Overall, {\bf \textit{Formulation 2 provides the greatest boost in mAP performance in the small dataset setting}}, with over $13\%$ mAP gain over the vanilla baseline and $50\%$ mAP gain over the deterministic baseline DKM. In general, all formulations provide similar benefits in the small dataset setting, with the most general performance benefit coming from Formulation 2. While Formulation 4 also provides a comparable performance boost, the performance difference from Formulation 1 and 2 may be attributed to compounding error from second-order approximation.
Interestingly, we note that modeling mean trajectories with kinematic priors is not enough to produce performance gain, as shown by the lower performance of DKM, which is deterministic with unit variance, and Formulation 4 with Learnable Covariance (DKM$_{Learnable}$ in Table~\ref{tb:1p_waymo}).

Compared to the results from Table~\ref{tb:100p_waymo}, the effects of kinematic priors in learning are more pronounced in smaller dataset settings.
Since kinematic priors analytically relate the position at one timestep to the position at the next, improvements in metrics may suggest that baseline models utilize a large amount of expressivity to model underlying kinematics.
In backpropagation, optimization of one position further into the time horizon directly influences predicted positions at earlier timesteps via the kinematic model. Without the kinematic prior, the relation between timesteps may be implicitly related through neural network parameters. 
When the model lacks data to form a good model of how an agent moves through space, the kinematic model can help to compensate by modeling simple constraints.

\subsection{Performance in the Presence of Noise}
We also show how kinematic priors can influence performance in the presence of noise. 
This is inspired by the scenario where sensors may have a small degree of noise associated with measurements dependent on various factors, such as weather, quality, interference, etc. or compounding error from other modules in the autonomy stack.

We evaluate the models from Table~\ref{tb:100p_waymo} when input trajectories are perturbed by standard normal noise $n_\epsilon \sim \mathcal{N}(0, 1)$; results for performance degradation are shown in Table~\ref{tb:perturb_waymo}. 

Table~\ref{tb:perturb_waymo} results are computed by measuring the \% of degradation of the perturbed evaluation from the corresponding original clean evaluation. We find that Formulation 4 from Section~\ref{sec:formulation4-accel-steering} with steering and acceleration components preserves the most performance in the presence of noise. This may be due to that second-order terms like acceleration are less influenced by perturbations on position, in addition to providing explicit constraints on vehicle movement. 
Additionally, distributions of acceleration are typically centered around zero regardless of how positions are distributed~\cite{9424171}, which may provide more stability in learning.

\begin{table}[t!]
 \caption{{\bf Percentage (\%) degradation of performance in the presence of noise. We also compare the robustness of each method by measuring the impact on performance in the presence of noise. In the table, we compute the relative change in performance between perturbed evaluation and clean evaluation, relative to each method. We observe that the model trained with {\em Formulation 4 (steering and acceleration representations)} offers the greatest robustness over other formulations in the presence of noisy inputs.}}
\label{tb:perturb_waymo}
  \centering
  \scalebox{.9}{
  \begin{tabular}{l|rrrr}
    \toprule
   Method &  mAP$\uparrow$ &  minADE$\downarrow$ &  minFDE$\downarrow$ &  MissRate$\downarrow$ \\
    \midrule
MTR & -2.965 & 2.302 & 0.671 & 1.928 \\
DKM$_{Learnable}$ & \textbf{-1.268} & 1.404 & 1.068 & 1.223 \\
Ours$_{F1}$ & -5.271 & 1.436 & 0.710 & 1.849 \\
Ours$_{F2}$ & -5.780 & 1.809 & 0.897 & 2.475 \\
Ours$_{F3}$ & -3.445 & \textbf{1.052} & \textbf{0.369} & 1.510 \\
Ours$_{F4}$ & -2.028 & 1.165 & 1.375 & \textbf{0.823} \\
    \bottomrule
  \end{tabular}}
  \vspace*{-1em}
\end{table}

%% file: sections/conclusion.tex
\section{Discussion and Conclusion}

In this paper, we present a simple method for including kinematic relationships in probabilistic trajectory forecasting. Kinematic priors can also be implemented for deterministic methods where linear approximations are not necessary. With nearly no additional overhead, we not only show improvement in models trained on robust datasets but also in suboptimal settings with small datasets and noisy trajectories, with up to {\bf $12\%$} improvement in smaller datasets and {\bf $1\%$ less} performance degradation in the presence of noise for the full Waymo dataset. \textbf{\textit{We find Formulation 2 with acceleration components to achieve the best rote performance, but Formulation 4 to be the most robust and fastest to converge.}} All experiments with kinematics modeled achieve non-trivial improved performance over the baseline model predicting trajectories directly.

When there is large-scale data to learn a good model of how vehicles move, we observe that the effects of kinematic priors are less pronounced. 
This is demonstrated by the less obvious improvements over the baseline in Table~\ref{tb:100p_waymo} compared to Table~\ref{tb:1p_waymo}; model complexity and dataset size will eventually out-scale the effects of the kinematic prior. With enough resources and high-quality data, trajectory forecasting models will learn to \textit{``reinvent the steering wheel"}, or implicitly learn how vehicles move via the complexity of the neural network.

One limitation is that we primarily explore analysis in one-shot prediction; future work focusing on kinematic priors for autoregressive approaches would be interesting in comparison to one-shot models with kinematic priors, especially since autoregressive approaches model predictions conditionally based on previous timesteps.

In future work, kinematic priors can be further explored for transfer learning between domains. While distributions of trajectories may change in scale and distribution depending on the environment, kinematic parameters, especially on the second order, will remain more constant between domains. 

\vspace*{1em}
\noindent
{\bf\large Acknowledgement: } This research is supported in part by IARPA HAYSTAC Program, Barry Mersky and Capital One E-Nnovate Endowed Professorships.

%% file: sections/appendix.tex
\newpage 
\onecolumn

\section{Appendix}

\subsection{Full Expansion of Formulation 3}
\label{sec:appendix_formulation3_expansion}
\begin{align*}
    \begin{bmatrix}
        \mathcal{N}_{x^{t+1}} \\
        \mathcal{N}_{y^{t+1}}
    \end{bmatrix} &= 
    \begin{bmatrix}
        (\mu^t_x + \sigma^t_x \cdot \epsilon_x) + (\mu^t_{s} + \sigma^t_s \cdot \epsilon_s) \cdot T_{\cos} (\mu^t_\theta + \sigma^t_\theta \cdot \epsilon_x) \cdot \Delta t \\
        (\mu^t_y + \sigma^t_y\cdot \epsilon_y) + (\mu^t_{s} + \sigma^t_s \cdot \epsilon_s) \cdot T_{\sin} (\mu^t_\theta + \sigma^t_\theta \cdot \epsilon_y) \cdot \Delta t
    \end{bmatrix} \\
    &= 
    \begin{bmatrix}
        (\mu^t_x + \sigma^t_x \cdot \epsilon_x) + (\cos(\mu_\theta) - \sin(\mu_\theta) \cdot \sigma_\theta \cdot \epsilon_\theta) \cdot (\mu_s + \sigma_s \cdot \epsilon_s) \cdot \Delta t \\
        (\mu^t_y + \sigma^t_y\cdot \epsilon_y) + (\sin(\mu_\theta) + \cos(\mu_\theta) \cdot \sigma_\theta \cdot \epsilon_\theta) \cdot (\mu_s + \sigma_s \cdot \epsilon_s) \cdot \Delta t 
    \end{bmatrix} \\
    &= 
    \begin{bmatrix}
        \begin{pmatrix}
            \mu^t_x  
            + (\mu_s \cdot \cos(\mu_\theta) \cdot \Delta t) \\
            + \sigma^t_x \cdot \epsilon_x 
            - (\mu_s \cdot \sigma_\theta \cdot \sin(\mu_\theta) \cdot \Delta t) \cdot \epsilon_\theta 
            + (\sigma_s \cdot \cos(\mu_\theta) \cdot \Delta t) \cdot \epsilon_s \\
            - (\sigma_s \cdot \sigma_\theta \cdot \sin(\mu_\theta) \cdot \Delta t) \cdot \epsilon_s \cdot \epsilon_\theta \\
        \end{pmatrix} \\
        \begin{pmatrix}
            \mu^t_y
            + (\mu_s \cdot \sin(\mu_\theta) \cdot \Delta t) \\
            + \sigma^t_y\cdot \epsilon_y 
            + (\mu_s \cdot \sigma_\theta \cdot \cos(\mu_\theta) \cdot \Delta t) \cdot \epsilon_\theta 
            + (\sigma_s \cdot \sin(\mu_\theta) \cdot \Delta t) \cdot \epsilon_s \\
            + (\sigma_s \cdot \sigma_\theta \cdot \cos(\mu_\theta) \cdot \Delta t) \cdot \epsilon_s \cdot \epsilon_\theta \\
        \end{pmatrix}
    \end{bmatrix} \\
\end{align*}

\subsection{Additional Results By Class}
In the paper, we present results on vehicles since we use kinematic models based on vehicles as priors. Here, we present the full results per-class for each experiment in Tables~\ref{tb:appendix-class-100p},~\ref{tb:appendix-class-1p},~\ref{tb:appendix-class-perturb-100p}, and ~\ref{tb:appendix-class-perturb-1p}. The results reported in the paper are starred (*), which are re-iterated below for full context. 

\begin{table*}[ht!]
 \caption{\textbf{Per-class results for performance on 100\% of the Waymo Dataset.}  } 
\label{tb:appendix-class-100p}
 \vspace{0.75em}
  \centering
  \scalebox{.8}{
  \begin{tabular}{c|c|rrrr}
\toprule
   Class & Method & $(\Delta\%)$ mAP$\uparrow$ & $(\Delta\%)$ minADE$\downarrow$ & $(\Delta\%)$ minFDE$\downarrow$ & $(\Delta\%)$ MissRate$\downarrow$ \\
    \midrule
\multirow{4}*{Average} & Baseline & 0 & 0 & 0 & 0 \\
& Ours + Formulation 1 & \textbf{1.7492} & -0.4455 & \textbf{-2.3882} & \textbf{-1.1098} \\
& Ours + Formulation 2 & -2.2235 & 0.1337 & -1.0881 & -0.7009 \\
& Ours + Formulation 3 & -0.9487 & \textbf{-0.4604} & -0.7560 & 1.2850 \\
\midrule 
\multirow{4}*{Vehicle*} & Baseline & 0 & 0 & 0 & 0 \\
& Ours + Formulation 1 & \textbf{2.376} & \textbf{-0.3444} & \textbf{-0.9102} & -0.3853 \\
& Ours + Formulation 2 & -0.2066 & 1.1069 & 0.1138 & -0.1651 \\
& Ours + Formulation 3 & -1.7045 & 0.246 & 1.0838 & 3.1921 \\
\midrule 
\multirow{4}*{Pedestrian} & Baseline & 0 & \textbf{0} & 0 & 0 \\
& Ours + Formulation 1 & \textbf{0.4657} & 0.2343 & \textbf{-0.7773} & -2.7692\\
& Ours + Formulation 2 &  -1.2806 & 0.885 & -0.2065 & \textbf{-3.8974} \\
& Ours + Formulation 3 & 0.0873 & 0.9630 & 0.6072 & -1.9487 \\
\midrule 
\multirow{4}*{Cyclist} & Baseline & 0 & 0 & 0 & 0 \\
& Ours + Formulation 1 & \textbf{2.4911} & -0.8991 & \textbf{-4.5646} & \textbf{-1.0656} \\
& Ours + Formulation 2 & -6.0854 & -1.1786 & -2.6418 & 0.1279 \\
& Ours + Formulation 3 & -1.2100 & \textbf{-1.8348} & -3.1439 & 1.0230 \\
    \bottomrule
  \end{tabular}}
\end{table*}

\begin{table}[H]
 \caption{\textbf{Per-class results for performance on 1\% of the Waymo Dataset.}  } 
   \label{tb:appendix-class-1p}
 \vspace{0.75em}
  \centering
  \scalebox{.8}{
  \begin{tabular}{c|c|rrrr}
\toprule
   Class & Method & $(\Delta\%)$ mAP$\uparrow$ & $(\Delta\%)$ minADE$\downarrow$ & $(\Delta\%)$ minFDE$\downarrow$ & $(\Delta\%)$ MissRate$\downarrow$ \\
    \midrule
\multirow{4}*{Average} & Baseline & 0 & 0 & 0 & 0 \\
& Ours + Formulation 1 & -1.6418 & \textbf{-6.3763} & \textbf{-14.9755} & \textbf{-5.173}\\
& Ours + Formulation 2 & \textbf{0.7463} & -0.7756 & -14.5275 & -2.9916 \\
& Ours + Formulation 3 & -6.1692 & 16.0354 & -11.2498 & -0.2493 \\
\midrule 
\multirow{4}*{Vehicle*} & Baseline & 0 & 0 & 0 & 0 \\
& Ours + Formulation 1 & \textbf{11.8444} & \textbf{-12.5280} & -27.1767 & \textbf{-8.3266} \\
& Ours + Formulation 2 & 6.7767 & -5.8432 & \textbf{-27.7645} & -7.2791 \\
& Ours + Formulation 3 & -5.3035 & 30.6413 & -20.5494 & -0.8327 \\
\midrule 
\multirow{4}*{Pedestrian} & Baseline & \textbf{0} & 0 & 0 & 0  \\
& Ours + Formulation 1 & -7.8373 & \textbf{-1.1325} & -1.2123 & 3.2820 \\
& Ours + Formulation 2 & -11.1275 & 1.7743 & -0.9999 & 6.2960 \\
& Ours + Formulation 3 & -3.2902 & -0.6795 & \textbf{-3.0440} & 0.1340 \\
\midrule 
\multirow{4}*{Cyclist} & Baseline & 0 & 0 & 0 & 0 \\
& Ours + Formulation 1 & -5.5283 & \textbf{-1.6251} & \textbf{-4.6731} & \textbf{-5.3754} \\
& Ours + Formulation 2 & \textbf{14.2506} & 3.8549 & -2.8120 & -2.4949 \\
& Ours + Formulation 3 & -11.9165 & 6.4701 & -2.5166 & 0.1588 \\
    \bottomrule
  \end{tabular}}
\end{table}

\begin{table}[H]
 \caption{\textbf{Per-class performance degradation results with perturbed evaluation for models trained on 100\% of the Waymo Dataset.}  }
 \label{tb:appendix-class-perturb-100p}
 \vspace{0.75em}
  \centering
  \scalebox{.8}{
  \begin{tabular}{c|c|rrrr}
\toprule
Class & Method & $(\Delta\%)$ mAP$\uparrow$ & $(\Delta\%)$ minADE$\downarrow$ & $(\Delta\%)$ minFDE$\downarrow$ & $(\Delta\%)$ MissRate$\downarrow$ \\
\midrule
\multirow{4}*{Average} & Baseline & \textbf{-2.5793} & 2.5097 & 0.7066 & 1.6939\\
& Ours + Formulation 1 & -2.9138 & 1.5662 & \textbf{0.3547} & \textbf{-0.6497} \\
& Ours + Formulation 2 &  -3.2141 & \textbf{1.4385} & 0.9715 & 1.8824 \\
& Ours + Formulation 3 & -3.5917 & 1.7306 & 0.7760 & 0.6344\\
\midrule 
\multirow{4}*{Vehicle*} & Baseline & -4.9587 & 4.7965 & 1.6527 & 3.5223\\
& Ours + Formulation 1 & \textbf{-4.9445} & \textbf{3.5542} & \textbf{1.4322} & \textbf{2.7072} \\
& Ours + Formulation 2 & -3.9596 & 2.4693 & 1.2800 & 2.7012 \\
& Ours + Formulation 3 & -4.7294 & 3.3984 & 1.3447 & 2.5600 \\
\midrule 
\multirow{4}*{Pedestrian} & Baseline & \textbf{-1.1932} & 0.8329 & 0.0243 & 3.3846 \\
& Ours + Formulation 1 & -1.5933 & \textbf{-0.2077} & \textbf{-0.7344} & \textbf{-1.0549} \\
& Ours + Formulation 2 & -3.066 & 0.4902 & 0.1582 & 2.7748 \\
& Ours + Formulation 3 & -2.0646 & 0.4383 & 0.4587 & 2.7197\\
\midrule 
\multirow{4}*{Cyclist} & Baseline & \textbf{-0.9253} & 1.0207 & 0.1255 & -0.5541\\
& Ours + Formulation 1 & -1.6667 & \textbf{0.4537} & \textbf{-0.1614} & \textbf{-3.0590}\\
& Ours + Formulation 2 & -2.3494 & 0.8361 & 1.0608 & 0.9366 \\
& Ours + Formulation 3 & -3.8905 & 0.6560 & 0.3594 & -1.7300 \\
    \bottomrule
  \end{tabular}}
\end{table}

\begin{table}[H]
 \caption{\textbf{Per-class performance degradation results with perturbed evaluation for models trained on 1\% of the Waymo Dataset.}  } 
 \label{tb:appendix-class-perturb-1p}
 \vspace{0.75em}
  \centering
  \scalebox{.8}{
  \begin{tabular}{c|c|rrrr}
\toprule
   Class & Method & $(\Delta\%)$ mAP$\uparrow$ & $(\Delta\%)$ minADE$\downarrow$ & $(\Delta\%)$ minFDE$\downarrow$ & $(\Delta\%)$ MissRate$\downarrow$ \\
    \midrule
\multirow{4}*{Average} & Baseline & -3.4328 & 0.5411 & 0.1187 & \textbf{0.1246} \\
& Ours + Formulation 1 & -6.2721 & \textbf{0.1349} & \textbf{-0.1171} & 0.6901 \\
& Ours + Formulation 2 & \textbf{-3.3086} & 0.7635 & 0.3181 & 1.574\\
& Ours + Formulation 3 & -4.0297 & 0.3420 & 0.8672 & 2.2805\\
\midrule 
\multirow{4}*{Vehicle} & Baseline & -6.0695	 & 1.1266 & \textbf{0.2207} & 0.9132\\
& Ours + Formulation 1 & \textbf{-4.4784} & 1.0241 & 0.8180 & 1.2892\\
& Ours + Formulation 2 & -5.6843 & 1.8236 & 1.3033 & 1.5933\\
& Ours + Formulation 3 & -5.6005 & \textbf{-0.3117} & 0.4686 & \textbf{0.4605} \\
\midrule 
\multirow{4}*{Pedestrian} & Baseline & \textbf{-0.2957} & \textbf{-1.1136} &\textbf{ -0.9734} & \textbf{-2.6122}\\
& Ours + Formulation 1 & -4.4124 & -0.3627 & -0.8599 & -0.0649 \\
& Ours + Formulation 2 & -3.0782 & -0.2411 & -0.3396 & 2.8355\\
& Ours + Formulation 3 & -4.8930 & -0.3421 & -0.2008 & 2.9431\\
\midrule 
\multirow{4}*{Cyclist} & Baseline & -5.8354 & 0.5518 & 0.4041 & \textbf{0.4083}\\
& Ours + Formulation 1 & -11.3134 & \textbf{-0.5455} & \textbf{-0.7410} & 0.5273\\
& Ours + Formulation 2 & -1.3978 & 0.1019 & -0.3669 & 1.1398\\
& Ours + Formulation 3 & \textbf{-0.6276} & 1.4908 & 1.6792 & 3.5779\\
    \bottomrule
  \end{tabular}}
\end{table}

\newpage 

\subsection{Experiment Hyperparameters}

\begin{table}[ht]
    \caption{\textbf{Model Architecture Hyperparameters}}
    \label{tab:appendix-model-architecture-hyperparamters}
    \centering
    \begin{tabular}{c|c|r} 
    \toprule
        Component & Hyperparameter & Value \\
        \midrule
        \multirow{4}*{Encoder} &  \# Hidden Features & 128 \\ 
         &  \# Attention Layers & 2 \\ 
         &  \# Attention Heads & 2\\ 
         &  Local Attention & True\\ 
         \midrule
         \multirow{4}*{Decoder} &  Hidden Features & 128\\ 
         &  \# Decoder Layers & 2\\ 
         &  \# Attention Heads & 2\\ 
         &  \# Hidden Map Features & 64\\ 
         \bottomrule
    \end{tabular}
\end{table}